\documentclass[10pt,twocolumn,letterpaper]{article}

\usepackage{iccv}
\usepackage{times}
\usepackage{epsfig}
\usepackage{graphicx}
\usepackage{amsmath}
\usepackage{amssymb}
\usepackage[T1]{fontenc}

\iccvfinalcopy 

\def\iccvPaperID{16} 

\ificcvfinal\fi

\begin{document}

\title{Weakly-Supervised Degree of Eye-Closeness Estimation}

\author{Eyasu Mequanint, Shuai Zhang, Bijan Forutanpour, Yingyong Qi, Ning Bi\\
Qualcomm AI Research 
\thanks{Qualcomm AI Research is an initiative of Qualcomm Technologies, Inc.} \\
San Diego, CA, USA. \\
{\tt\small \{emequani, shuazhan, bijanf, yingyong, nbi\}@qti.qualcomm.com}}

\maketitle
\ificcvfinal\thispagestyle{empty}\fi
\begin{abstract}

Following recent technological advances there is a growing interest in building non-intrusive methods that help us communicate with computing devices. In this regard, accurate information from eye is a promising input medium between a user and computing devices. In this paper we propose a method that captures the degree of eye closeness. Although many methods exist for detection of eyelid openness, they are inherently unable to satisfactorily perform in real world applications. Detailed eye state estimation is more important, in extracting meaningful information, than estimating whether eyes are open or closed. However, learning reliable eye state estimator requires accurate annotations which is cost prohibitive. In this work, we leverage synthetic face images which can be generated via computer graphics rendering techniques and automatically annotated with different levels of eye openness. These synthesized training data images, however, have a domain shift from real-world data. To alleviate this issue, we propose a weakly-supervised method which utilizes the accurate annotation from the synthetic data set, to learn accurate degree of eye openness, and the weakly labeled (open or closed) real world eye data set to control the domain shift. We introduce a data set of 1.3M synthetic face images with detail eye openness and eye gaze information, and 21k real-world images with open/closed annotation. The dataset will be released online upon acceptance. Extensive experiments validate the effectiveness of the proposed approach.

\end{abstract}
\vspace{-0.75em}
\section{Introduction}
\label{sec:intro}

Several advanced input technologies have been proposed to simplify user's interactions with computing devices. Information from an eye is one of the input techniques which improves the experience of working with a computer. While measuring a user's visual line of gaze (where s/he is looking in space) has been
improving, the degree of eye closeness - which is rich in information for applications such as user-computer dialogue \cite{krolak2012eye} - has not been well studied. Detection of human eyelid openness or blink state is a key step for effective eye-based vision systems. There are plenty of applications that require accurate eye states estimator: user-computer interaction \cite{krolak2012eye}, face authentication system where eye states are used for user's attention assessment and anti-spoofing \cite{pan2007eyeblink,szwoch2012eye, oh2012timing}, photography, deceit detection \cite{marchak2013detecting,peth2013fixations}, emotion analysis, eye tracking, avatar animation, gaming, virtual reality, and driver's drowsiness detection which help avoid impairment that leads to, according to The American National Highway Traffic Safety Administration (NHTSA), $\$109$ billion in damages annually. 

Typically, a computer vision approach for eye states detection first extracts features around eyes part, and then classify the eye states. Existing approaches are designed only for binary eye states detection (open or closed). In an eye blink detection system, the process needs to collect multiple image frames as input. However, due to the speed of an eye blink, a fully eye-closed image may not be captured/sampled, and thus a binary state system could easily lead to incorrect decision due to missed blinks. A binary state is often insufficient for accurate user-computer dialogue and other similar applications that require higher speed and accuracy, such as photography, anti-spoofing, and others. For example, one of the goals of advanced driving assistance systems is early detection of driver's drowsiness. An alert is raised when the eyes are at least $80\%$ closed over a certain time period \cite{Dinges_Grace_1998}. As in prior examples, a two-state eye openness system is insufficient for high accuracy. Estimation of eyelid openness with more granularity allows for the extraction of more meaningful information for addressing these real-world applications. An example of detailed and binary eye states is shown in Figure \ref{fig:face_degree_plot}.

In this work, we develop a deep neural network (DNN) based framework that can detect the degree of eye-openness with high granularity. It provides more accurate and detailed information than current binary states (open/closed) systems. Using deep learning for eye openness requires highly granular and accurately annotated training data. Such training data is often scarce and cost prohibitive. To address this problem, we introduce a large data set of synthetic face images rendered using advanced graphics techniques with accurately controlled degree of eye openness (Figure \ref{fig:face_degree_plot}), and a limited set of real face images with binary eye states labels. One issue that arises is the domain shift between these synthetically generated data vs authentic real world face images. To overcome this, we propose a weakly-supervised training method which utilizes the accurate annotation from the synthetic data, and weak annotation (open or closed) on recorded data for eye openness estimation. The contribution of our work is listed as follows:

\begin{figure*}
\begin{center}
\includegraphics[trim=0cm 0cm 0cm 0cm, width=.85\linewidth]{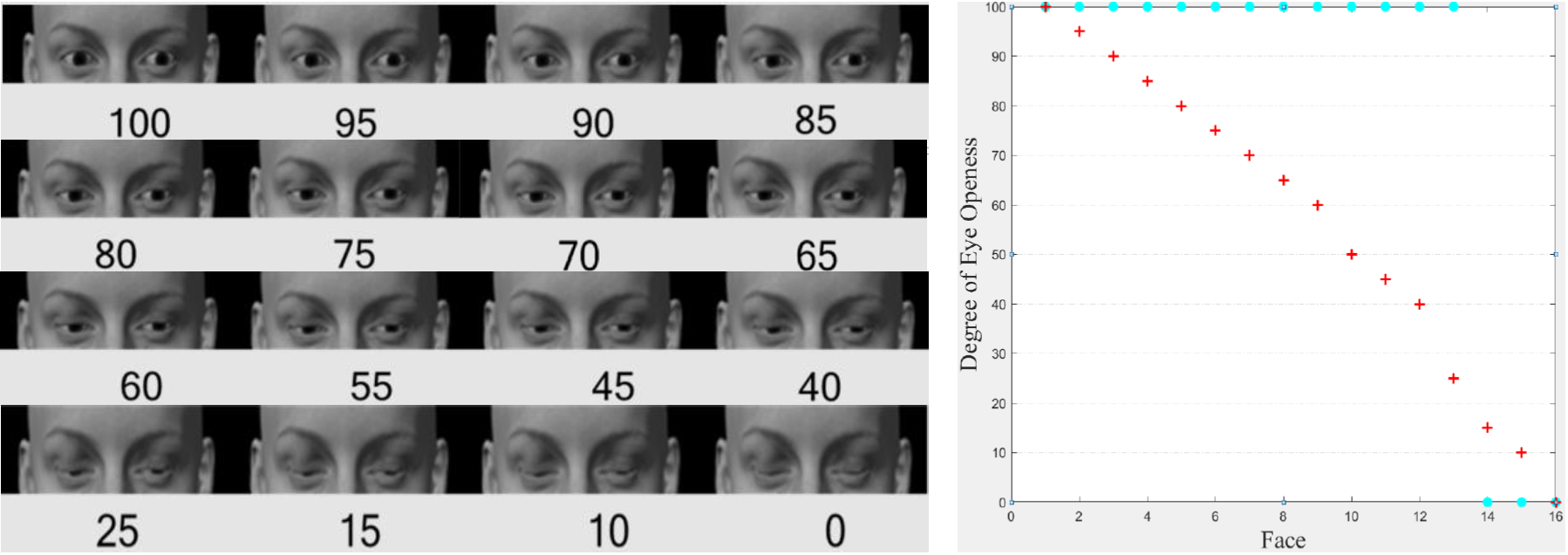}
\end{center}
   \caption{\small{{\bf{Left:}} eye portion of synthetic faces with labeled degree of openness (100 and 0 refers fully open and fully closed respectively). {\bf{Right:}} cyan simulates results of available eye openness detection and red simulates results from our proposed approach.}}
\label{fig:face_degree_plot}
\end{figure*}

\begin{itemize}
    \item A computer vision based system to detect eye openness with high granularity for several applications such as human computer interaction. Our approach achieves high granularity results from low-granularity, binary labeled (opened or closed eye) real-world images.
    \item Augmenting, using weakly-supervised learning, the real-world training images with binary annotation (opened or closed eye) with synthetically generated images with detail information on the degree of eye openness. We introduce 1.3M synthetic face images (Figure \ref{fig:face_degree_plot}) and 21K real-world images.
    \item We conduct experiments which show that the proposed approach effectively estimates the degree of eye openness for real-world image with high accuracy and granularity.
\end{itemize}

\section{Related Works}
Several eye-based systems have been proposed in the literature which use the percent of closeness (PERCLOSE) and average eye closure speed (AECS) measures for different decisions, such as drowsiness detection where PERCLOSE increases \cite{darshana2014efficient,daza2011drowsiness,rezaee2013real,sheng2013integrated,manu2016facial,dasgupta2013vision,pratama2017review,danisman2010drowsy, diana2018in} and AECS decreases \cite{fors2011camera, barr2005review, bergasa2006real}, for a drowsy driver. Existing eye-based approaches mostly use eye and face detectors, such as Viola Jones algorithm \cite{viola2004robust}, and detect the eye state using classical computer vision techniques. \cite{darshana2014efficient} trained a Support Vector Machine (SVM) for eye state classification. Tomas \etal{} \cite{drutarovsky2014eye} divided the eye region into 3$\times$3 cells where local motion vectors are estimated whose variance of the vertical components is used to determine the eye state. \cite{daza2011drowsiness} detected the eye states equalizing the eyes using a Hat transformation followed by eye tracking strategy in a sequence of frames. \cite{pan2007eyeblink} introduced an appearance based image feature to detect the eye openness using the AdaBoost algorithm. \cite{jimenez2013optical} detect the eye states by analyzing the response of a horizontal Laplacian filter around the eyes; numerous vertical line segments should be visible, due to the pupils and eye corners, when the eyes are open, and only horizontal lines are observed when eyes are closed. \cite{lee2010blink} detects the closed and open eye states based on the number of black pixels the eye has; in a binarized eye region, closed eye image has higher number of black pixels compared to an open eye image. \cite{sukno2009automatic} first detects $98$ facial landmarks and the average height-width eye ratio is used to determine the eye's state in a given frame.  A real-time eye state detector, designed for very low near-infrared image, is proposed in \cite{marc2007Real}. More recently, \cite{geng2019real} introduced deep learning into the field of fatigue detection. The method detects the face and feature point locations using multi-task cascaded convolutional neural network (MTCNN). The eye region is obtained according to the geometric relationship between eye feature points, and then eye state is classified by convolutional neural network (CNN). 

\begin{figure*}[t]
\begin{center}
\includegraphics[trim=0cm 0cm 0cm 0cm, width=0.9\linewidth]{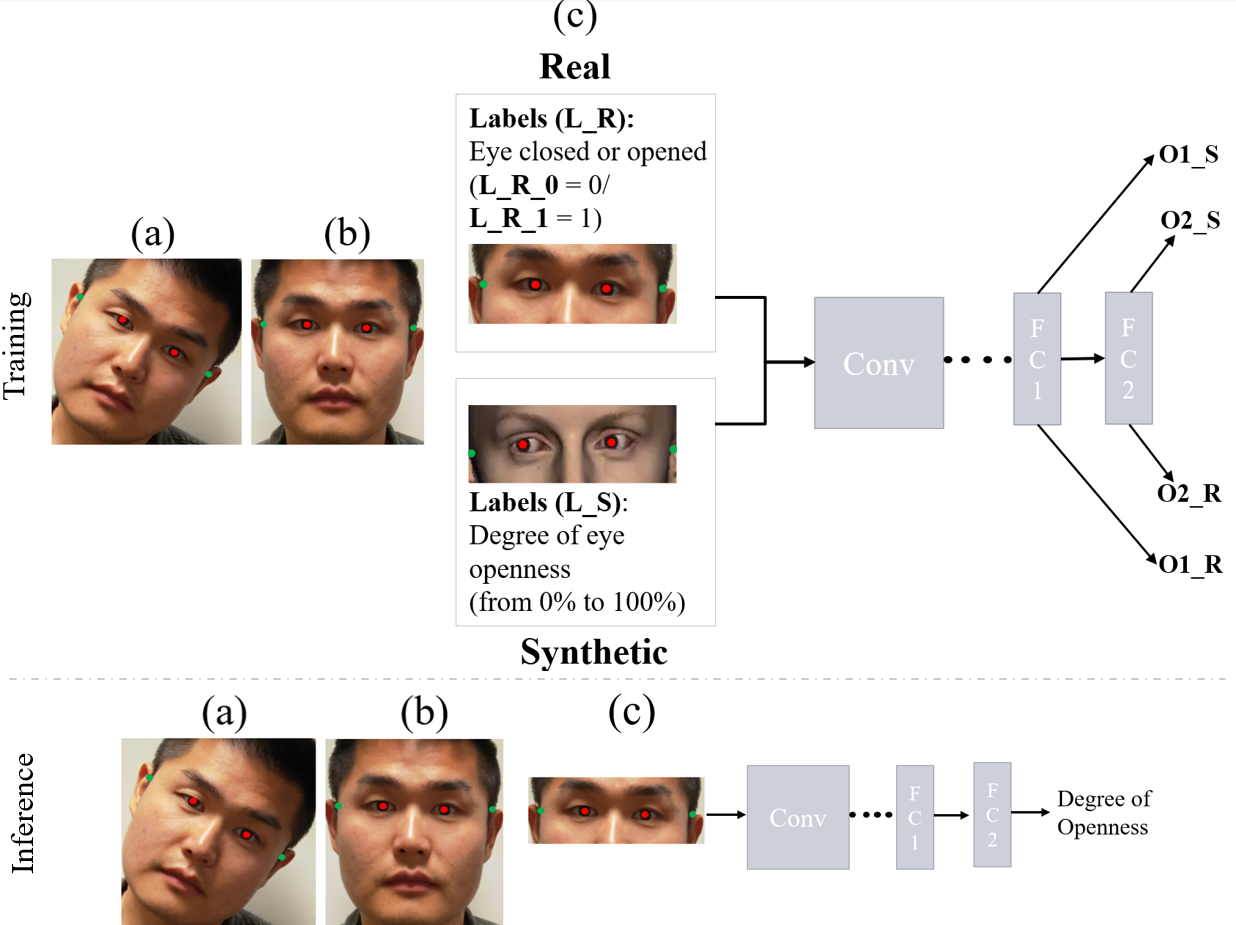}
\end{center}
   \caption{ \small{The proposed architecture for estimating degree of eye openness (training upper and inference bottom). (a) raw face image with landmarks (could be real or synthetic), (b) normalized version of the face, and (c) last stage of the preprocessing, contains cropped eye portion of the face. The preprocessed real and synthetic data are separated into two different groups with their corresponding labels ($\mathbf{L\_R}$ and $\mathbf{L\_S}$). `Conv', `FC1' and `FC2' represent  a shared convolution block, and two fully connected blocks respectively. The output of FC2 estimates the degree of eye openness. $\mathbf{O1\_S}$ = output of the synthetic data at FC1, $\mathbf{O1\_R}$ = output of the real data at FC1, $\mathbf{O2\_S}$ = output of the synthetic data at FC2, $\mathbf{O2\_R}$ =  output of the real data at FC2. $\mathbf{O2}$ is a scalar which represents openness amount and $\mathbf{O1}$ is feature vector of size 256.} }
\label{fig:network}
\end{figure*}

All the above referenced works and other several eye state based systems detect only two levels of eye opening, which is not enough to model a practical system. From a perspective of practical real world system applicability, the eye state detection system should satisfy several constraints which could not be solved just by using only the two eye states. Drowsiness is a very good practical example which is a state a driver might be in with a partially closed eye. Very few works introduced percentage of eye openness which is more accurate than methods that detects only two levels of openness \cite{anas2017online, han2015driver, akrout2015spatio}.  \cite{anas2017online} just added a third level (partially opened) to the two eye states (opened and closed). \cite{akrout2015spatio} and \cite{han2015driver} are drowsiness detection methods that uses the notion of percentage of eye openness (various states of eye openness). Both of them, to detect detailed eye states, use classical computer vision techniques. \cite{akrout2015spatio} is a geometry shape-based approach which uses Circular Hough Transform method to localize iris and eyelids. Since it's a geometric-based approach, a very small variation in eyelids localization leads to a wrong decision, and it also easily gets affected by illumination variation. Our approach, since we use deep learning, does not need iris and eyelids detection and is more robust for illumination variations. \cite{han2015driver} is a video-based solution for eyelids movement detection. Classical approaches are used to detect the face and eye and only the left eye part is then taken as input to the next stage which vectorizes the input, does dimension reduction and input the result to a single linear model which detects the eye openness score. The eye openness score is then passed to a clustering module which helps the system detect some pattern based on which eyelids movement is detected. The method is not a general model, not fully automated and is subject dependent, and the structure of features extraction scheme needs to be defined by the user. It needs different levels of feature clustering whose number of clusters should be known for different level of feature extractions. The system, in general, has 6 parameters to tune which make it harder to be used in fully automated systems.

\section{Proposed approach}

We propose a deep learning solution to estimate the degree of eye openness. 
One of the things which make solving this problem using deep learning difficult is data. The high performance of deep learning results from abundant labelled training data. Collecting a data with accurate degree of eye openness is a difficult task. To best serve our purpose we used a synthetic data with different levels of eye openness. Using synthetically generated data, though solves the data scarcity, has a problem in training a general-purpose deep learning model for our task. Models trained on the synthetic data, due to the domain gap the data has with the real data, fail when tested on a real dataset. Domain adaptation studies the domain shift problem for the better use of available training data for new testing domains. In this work we propose to train a model, using a synthetic data that have known levels of eye openness and real data that only have open/closed annotation,  which help us get the different eye states on real test dataset addressing the domain shift our data has with the synthetic data. 

During training, the input batch to the network contains images from both the synthetic and real ones. The network, as shown in figure \ref{fig:network}, consists of an input which comes from both synthetic and real data blocks, a convolution block (Conv) and two fully connected blocks (FC1 and FC2). Given an image, in the first step we detect the face and landmarks utilizing dlib toolkit \cite{dlib09}. In step (b) we align and normalize the face based on the landmarks shown by red and green dots, and the eye portion of the face is cropped to be passed through the newtwork. The output of the FC2 block regresses the degree of eye openness (0 means closed, and positive numbers represent different levels of eye openness). The real data has no detail eye openness annotation. It just has information if the eye is closed or open, a kind of dataset much cheaper than a dataset with labeled degree of eye openness. We train the network leveraging all the available information both from the synthetic and real data. To this end, we propose a loss combined from three different losses.

\begin{figure*}[h!]
\centering
\includegraphics[trim=0cm 0cm 0cm 0cm, width=0.95 \linewidth]{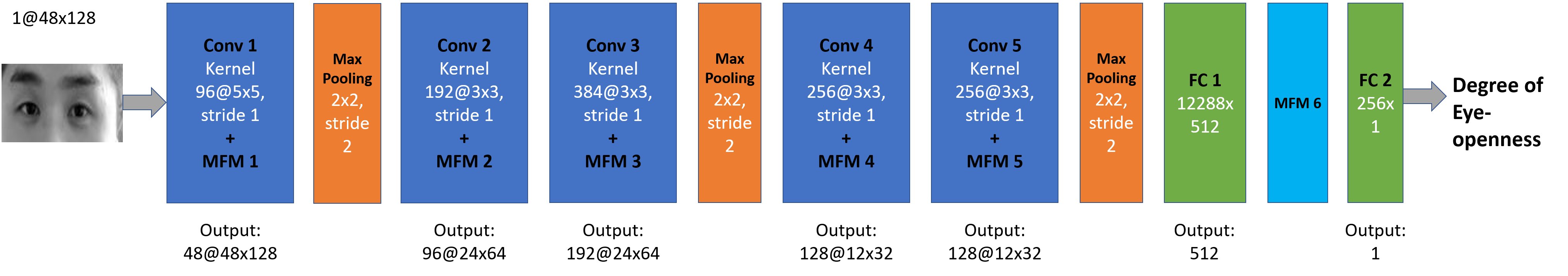}
   \caption{\small{Model architecture. }}
\label{fig:model arch}
\end{figure*}

{\bf Architecture:} Very light network architecture based on Max-Feature-Map (MFM) operation, neural inhibition operation proposed  in \cite{wu2018light}, is used. MFM operation is a special case
of maxout \cite{goodfellow2013maxout} to learn a light convolutional neural network (CNN) with a small number of parameters. It is an alternative of ReLU which adopts a competitive relationship
to suppress low-activation neurons in each layer. It not only is able to separate noisy and informative signals but also does feature selection between two feature maps. We refer the reader to \cite{wu2018light} for better understanding of the MFM operation. 'Conv' (Figure \ref{fig:network}) is constructed by 5 convolution layers with Max-Feature-Map operations and 3 max-pooling layers, shown in Figure \ref{fig:model arch}. The 256-D deep features are extracted from the output of fully connected layer after MFM operation (layer MFM 6).

{\bf Loss Functions:} The main challenge for training our model is the lack of ground truth for the real data. To address the problem we use a combination of three losses which leverages recent ideas from the problem of domain adaptation \cite{Sankaranarayanan_2018_CVPR}. 

First, mean square error (MSE) loss is used, for inputs from the synthetic data, to reduce the gap between the regressed degree of eye openness and the ground truth labels, see \textit{Loss1} in the follow paragraph. This loss is the key part to facilitate the proposed weakly supervised training, it relieve the painful detailed granular eye openness annotations for real-world dataset. 

The second loss uses information we have for the real data, eye is closed or open, please see \textit{Loss2}. If the real input data is labelled as closed, the output prediction from our network should be zero, otherwise the network should output a number greater than a predefined threshold, openness threshold ($\mathbf{OT}$). $\mathbf{OT}$ = Openness threshold is a hyper parameter, used with the formula:
$$
    \begin{cases}
        1 \textit{ (open eyes)},  \textit{  if } \mathbf{O2\_R} > \mathbf{OT}; \\
        0 \textit{ (closed eyes)},  \textit{ Otherwise};
    \end{cases}
$$

Recent approaches which train a network to bring the source and target distribution together show excellent performance \cite{Sankaranarayanan_2018_CVPR}. This inspired us to introduce a distribution loss, a loss which help us train our network using the gradients from the change in the distribution from the synthetic and real input data, see \textit{Loss3}. This helps us bring the source
and target distributions closer in the feature space learned by the network.

The proposed loss functions for cross domain (synthetic and real) network training consists of the following three terms:

\begin{itemize}
    \item \textbf{Loss1}: MSE loss for accurately predicting level of openness on synthetic data \\ \\
            \hspace*{0.5cm}    \textit{Loss1} = MSE\Big($\mathbf{O2\_S}$, $\mathbf{L\_S}$\Big), 
    
    where $\mathbf{O2\_S}$ is the estimated eye degree of synthetic data and $\mathbf{L\_S}$ is the accurately labeled eye degree, see Fig 2;
            
    \item \textbf{Loss2}: Binary loss for accurately predicting binary (opened/closed eye) labelled real data with $\mathbf{OT}$  \\
            \hspace*{0.5cm}\textit{Loss2} = $\frac{1}{N}$ $ \sum_{i}$ $\Big \{||\mathbf{O2\_R}_i||^2*(1-\mathbf{L\_R_i}) + max \Big((\mathbf{OT}-\mathbf{O2\_R}_i), 0\Big)*\mathbf{L\_R_i}\Big \}$, 
            
    where $\mathbf{O2\_R}$ is the estimated eye degree of real image data and $\mathbf{L\_R}$ are the corresponding binary open/close labels. $\mathbf{OT}$ is the eye openness threshold;
            
    \item \textbf{Loss3}: Distribution loss for controlling domain mismatch between synthetic and real data;  the distribution from synthetic and real data should be similar. \\ \\
            \hspace*{0.5cm}    \textit{Loss3} = $abs\Big(mean(\mathbf{O1\_S})-mean(\mathbf{O1\_R})\Big)+  abs\Big(var(\mathbf{O1\_S})-var(\mathbf{O1\_R})\Big)$,
            
    where $\mathbf{O1\_S}$ and $\mathbf{O1\_R}$ are the feature vectors of synthetic/real data from the `FC1' layer of the model, see Fig 2. 
    
\end{itemize}
The final loss is computed as 
$$
\textit{Loss} = \lambda_1\textit{Loss1} + \lambda_2\textit{Loss2} + \lambda_3\textit{Loss3}.
$$

\section{Q\_ECE: Eye Openness Estimation Dataset} 

\begin{figure*}[t]
\centering
    \includegraphics[clip, trim=1cm 0cm 1cm 0cm, width=.85\linewidth]{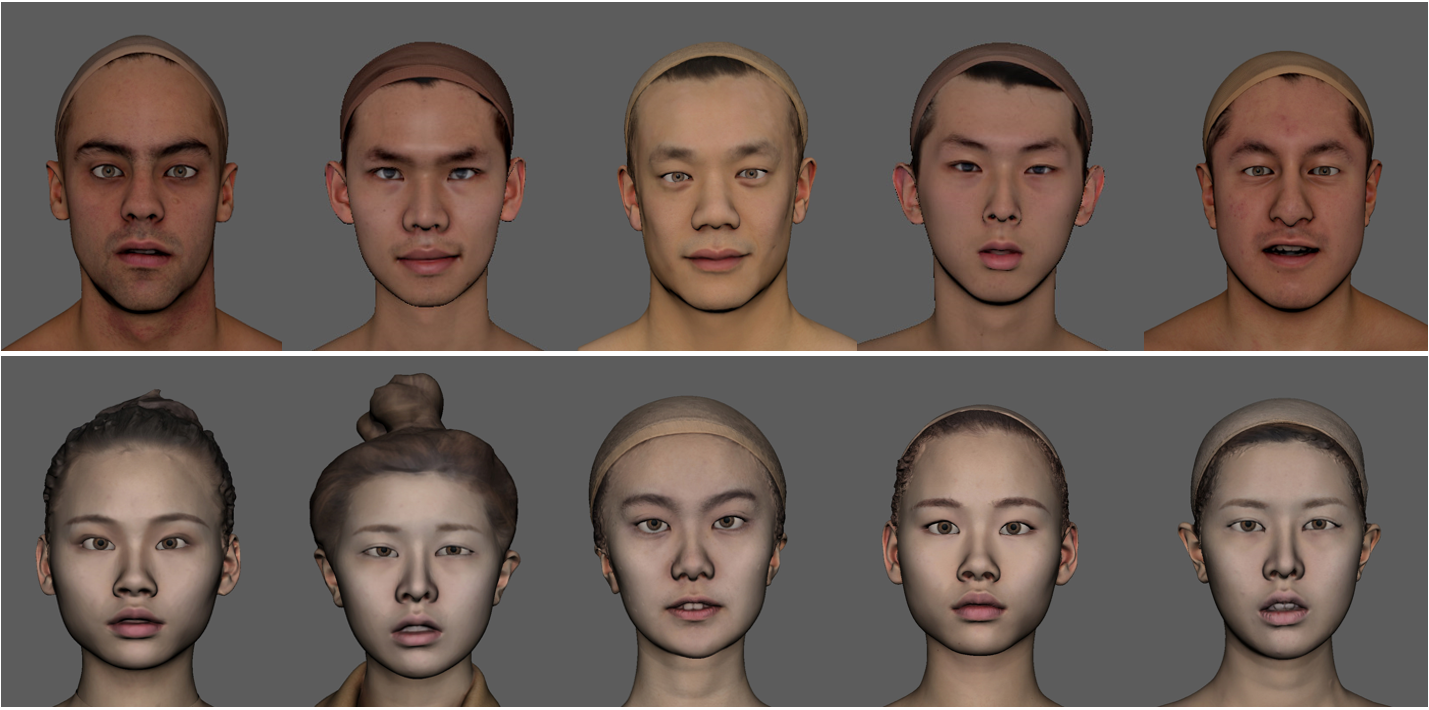}
   \caption{\small{Sample synthetic face models.}}
\label{fig:face_models}
\end{figure*}

Although problems related to eye-based systems have received a lot of attention, their performance still has not reached an acceptable level for practical use cases.  The lack of high quality training data is one of the issues which limits the development of such systems. Existing databases, such as ZJU \cite{pan2007eyeblink}, Eyeblink8 \cite{drutarovsky2014eye} and Silesian5 \cite{radlak2015silesian} lack the information necessary to address essential challenges. The datasets mentioned above do not take into account important characteristics such as human pose or image illumination.  Furthermore, the samples are captured from a limited number of subjects.  Another requirement of high quality datasets is high quality annotation of the data. In order to work well in real-world conditions, most eye-based systems require knowledge of all possible eye states.  One of the main reasons existing eye-based approaches do not perform well for real-world applications is the low number of `eye states' they use. This is typically only two: closed or open. Many existing real-world data sets require additional annotation, which can be cost prohibitive. 

To alleviate the data annotation burden we create a dataset of 1.3M synthetic data by rendering face images using computer graphics techniques.  The dataset was created using high quality 3D scans of 13 human head models 6 female and 7 male from different age groups and ethnicities \footnote{The 3D scans could be found here: www.3dscanstore.com}. Eyeball models were separate, and placed in the 3D eye sockets. Additionally, the subjects' eyelids were animated in conjunction with the up and down rotation of the eyeball. The dataset included 198 eye directions, via look-at target points (11 vertical x 18 horizontal, in a grid pattern), +/- 25 degrees vertical, +/- 35 degrees horizontal, in 5 degree increments. In addition, there were 11 different states of eye openings (100 \% open to close in 10 \% increments). Finally 49 camera positions were generated (-30 to 30 degree in 10 degree increments, horizontally and vertically). Some examplar images from this synthetic dataset are shown in Figure \ref{fig:face_models}. Our work is the first to create and use such a large dataset of high quality rendered face images with controlled eyelid movement. Since the dataset contains eyeball rotation and 2D and 3D look at point information as well, the data is not only useful for eye openness estimation, but also for gaze estimation and attention detection.

In addition to the synthetic data we also collected a real data from 16 subjects, different age groups and ethnicity,  using NIR and RGB sensors. In the real data collection we tried to consider several situations as pose, illumination, (sun) glasses and others. For the case of pose variations we asked the subject to move the head 360 degree and in four different directions (Left, right, up and down). We consider the illumination variation collecting the data from different environments as 'indoor full light', 'indoor low light', 'in a dark' (NIR sensor only), 'outdoor shade' and 'outdoor sunlight'. '(Sun) glass' and 'No (sun) glass' situations are also included in most of the subject's data. In total we collected around 21k real images (17k NIR and 4k  RGB images) with more than 12k closed and around 9k opened eyes. We also collected, from four of the subjects' a data which covers detail eye states, we asked the subjects to close and open the eye very slowly. A good model which estimates detail eye openness should give us 'U' kind of shape on plotting the frames versus degree of eye openness, please see the result on figure \ref{fig:close-open-close-open}    

\section{Experiments and results}
Since, to the best of our knowledge, there is no available dataset and recently proposed methods related to detailed eye states estimation, we conduct experiments on Q\_ECE dataset introduced in this work. 
\subsection{Implementation Details}
We implemented our method with Pytorch. For all training we take 80 epochs with an initial learning rate of 0.0001 and a batch size of 256. In case of joint training the percentage of input data (from a batch) is fixed as 25\% real and 75\% synthetic. Based on the generated synthetic data, as shown in figure \ref{fig:face_degree_plot}, and using some experimental validation, we observed that a openness threshold ($\mathbf{OT}$) of 15 gives the best result. $\lambda_1$ is set to 0.01 and both $\lambda_2$ and $\lambda_3$ are set to 1. For all experiments with synthetic data, we learn eye openness estimation using the synthetic data branch only. For the real-world datasets we use a joint training with the synthetic dataset, any input batch to the network contains images from both synthetic and real images. For an effective transfer learn from synthetic data, we follow similar ideas from existing works \cite{wood2015rendering, zhang2019mpiigaze}; it is important that both the synthetic and real dataset have the same distribution of eye openness. In our model, gray-scale face images are used instead of RGB images. The face images are aligned to 144$\times$144 by the landmarks and the eye portion of 48$\times$128 is cropped and used as inputs to the 'Conv' layer. Besides, each pixel value is normalized to be between [0, 255].

\subsection{Results and Metrics}

\begin{figure*}[h!]
\begin{center}
\includegraphics[trim=0cm 0cm 0cm 0cm, width=.8\linewidth]{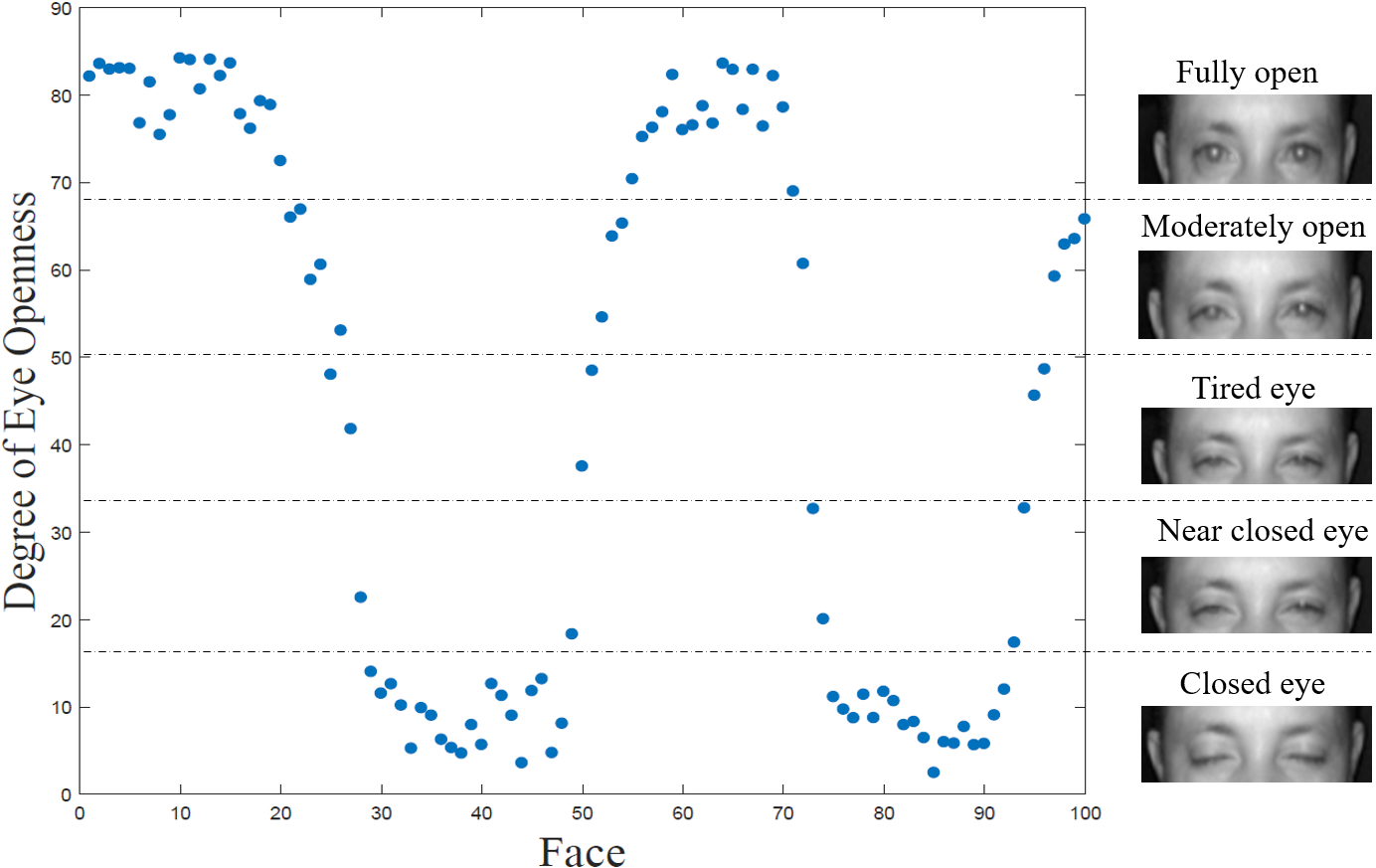}
\end{center}
   \caption{\small{Performance on (100) video frames captured with "Close-Open-Close-Open" sequence moving the eyelids very slowly.} }
\label{fig:close-open-close-open}
\end{figure*}

\begin{figure*}[h!]
\begin{center}
\includegraphics[trim=0cm 0cm 0cm 0.5cm,width=.8\linewidth]{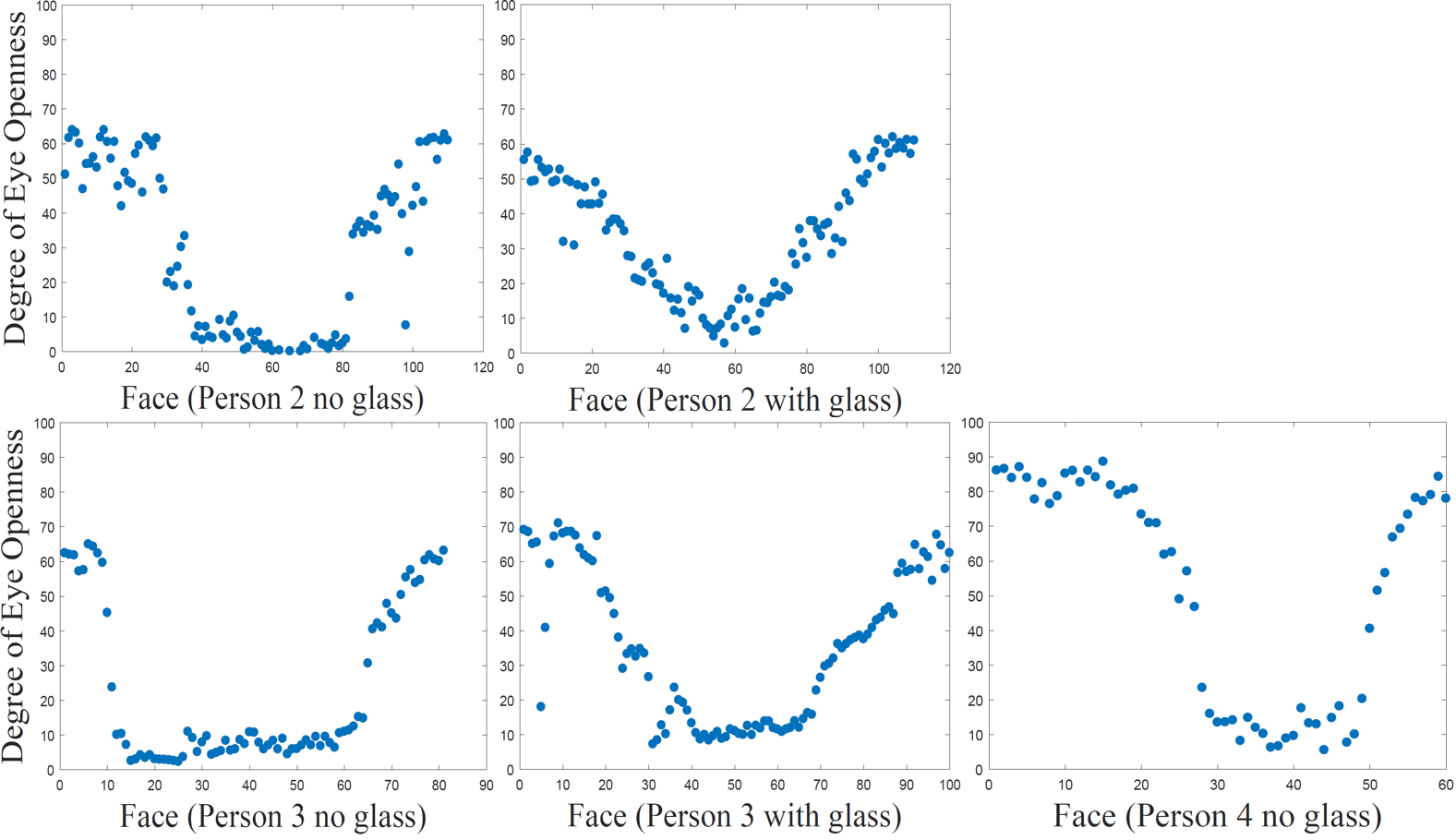}
\end{center}
   \caption{\small{Performance on a video with "Close-Open" sequence moving the eyelids very slowly.}}
\label{fig:close-open-close-open-3-person}
\end{figure*}

The  performance of degree of eye openness is measured using the mean squared error (MSE) computed between the regressed degree of eye openness and the ground truth eye states. We believe that MSE is the best metric for our purpose (detail eye openness estimation). When we label the dataset, to avoid the resolution and the camera distance issues, we consider the normalized face image based on detected landmarks; a person whose image is taken from different camera distances and different sensors with different resolution should have similar openness amount for similar eye states. The openness amount from our data ranges  from 0 (fully closed) to 100 (fully open). 100 (fully open eye state) is given to the biggest eye from all eyes in our dataset. That means if we test using a bigger eye than the biggest eye from our dataset, the eye openness estimation result will be beyond 100, and openness amount of a fully open small eye will be much smaller than 100. The performance of eye openness and closeness is measured using accuracy metric. 

\begin{table}[h!]
  \begin{center}
   \begin{small}
    \begin{tabular}{l|c|c|c} 
      \textbf{Training data} & \textbf{Test data} & \textbf{Degree of eye } & \textbf{Open/Close} \\
         &  &  \textbf{openness (MSE)} & \textbf{(Accuracy)} \\
      \hline
      Synthetic & Synthetic & 9 & 100\\ \hline
      Synthetic  & Real & -- & 47.5\\ \hline
      Synthetic + Real & Synthetic & 9 & 100\\ \hline
      Synthetic + Real & Real & -- & 99.62\\ \hline
      Real & Synthetic & 8094.8 & 52.3\\ \hline
      Real & Real & -- & 96.30\\ \hline
    \end{tabular}
   \end{small}
   \end{center}
    \caption{\small{Results on Q\_ECE dataset, with varying training source}}
   \label{tab:table1}
\end{table}


Table \ref{tab:table1} shows quantitative results for the various input configurations of our eye openness estimation network. The degree of eye openness, evaluated using MSE metric, is same with the two input configurations. The result tells us that the degree of eye openness for the synthetic test data deviates, on average, from the ground truth only by 3\%. Considering 8\% variations, in degree of eye openness, from the ground truth annotation as correct eye openness estimation we end up with 100 \% accuracy in estimating the openness or closeness of the eyes. We observe that providing the real dataset together with the synthetic as input to our network (for joint training) results improved accuracy of eye openness and closeness detection on the real data. Training only using one (real or synthetic) and testing on the other end up with a result close to random (open/close) decision. It is also noticed that the joint training, since it helps to augment training samples and regularize model from overfitting, improves the accuracy on the real-real experiment scenario. 

We also evaluated our model by assessing its performance on video sequences collected from four of the subjects which we have asked to close and open their eye with very slow motion which help us have various eye states.  Figure \ref{fig:close-open-close-open} shows an example which is plotted from the predicted degree of eye openness from the video frames. A blue point represents the degree of eye openness (y-axis) of a face (x-axis). The entire process of closing and opening the eye could be represented by a curve with `U' kind of shape. This help us extract meaningful information which could be leveraged for different applications. As can be seen from the figure, our model was able to capture the detail eye states, and we could divide the states (based on the degree of openness on the y-axis) into different meaningful information as `fully open', `moderately open', `tired eye', `near closed eye' and `closed eye'.

Experimental results on the other three subjects whose data is collected with "Close-Open" sequence moving the eyelids very slowly is shown on figure \ref{fig:close-open-close-open-3-person}. As shown from the figure the proposed framework is able to capture detail eye openness states for all the persons with and without glass.  


The last experiment which we conducted is using the small subset of our real dataset that are annotated with detail eye openness, let us call it Real'. Real' consists of 2000 images. We first compute the red and green points as shown on figure \ref{fig:network} and used them to align and normalize the face and then we annotate the upper and lower eyelid points which is used to compute degree of eye openness that is used as our ground truth. 75\% of the data is used to fine-tune the model trained using 'Synthetic + Real' and the rest of the data is used for testing. During training, since in this case we have detail eye openness annotations, we compute the loss for this dataset as losses that consider synthetic data annotations. The performance is then measured using 'MSE' metric. As shown on Table \ref{tab:table2} the joint training help us boost the test result on real-world images. Moreover, adding few samples with detail annotations has a minor improvement over the binary labelled real data.

\begin{table}[!ht]
\begin{small}
\centering
    \begin{tabular}{l|c|c} 
      \textbf{Training data} & \textbf{Test data} & \textbf{Degree of eye } \\
         &  &  \textbf{openness (MSE) } \\
      \hline
      Synthetic & Real' (test) & 2045.80 \\ \hline
      Synthetic + Real & Real' (test) & 45.35 \\ \hline
      Synthetic + Real + Real' (train) & Real' (test) & 34.20 \\ \hline
    \end{tabular}
    \caption{\small{Result on Real' face database}}
   \label{tab:table2}
\end{small}
\end{table}

\section{Conclusion}
In this work, we shed the light to the research field of degree of eye openness estimation which help us estimate detail eye states, a problem which has not been well studied. We have addressed essential issues of the problem in-terms of practical and theoretical contributions. First, we created fully annotated synthetic data for estimation of the degree of eye openness which  release the burden of detail eye state annotation of real images. The dataset will be released online upon acceptance. Secondly, we introduce a  weakly-supervised problem  of leveraging low-cost binary labelled (opened or closed eye) real images together with the synthetic data for accurate estimation of the degree of eye openness on the real images. To this end we also collected real data which considers different practical situations. The experiments verify that the proposed approach effectively estimates the degree of eye openness for real world image. The proposed method, leveraging the cheap synthetic images, adapt easily to a weakly labelled real-world images.

{\small
\bibliographystyle{ieee}
\bibliography{egbib}
}

\end{document}